# EXPECTATION-MAXIMIZATION TECHNIQUE AND SPATIAL - ADAPTATION APPLIED TO PEL-RECURSIVE MOTION ESTIMATION

**Vania Vieira Estrela**
Universidade Estadual do Norte Fluminense, LCMAT-CCT, Grupo de Computação Científica
Av. Alberto Lamego 2000, Campos dos Goytacazes, RJ, BRAZIL, CEP 28013-600
E-mail: vaniave@uenf.br

**M. H. da Silva Bassani**
Departamento de Engenharia Mecânica, DEPES, CEFET-RJ
Av. Maracanã 229, Rio de Janeiro, RJ, BRAZIL, CEP 20271-110
E-mail: marcos@cefet-rj.br

**ABSTRACT**

Pel-recursive motion estimation is a well-established approach. However, in the presence of noise, it becomes an ill-posed problem that requires regularization. In this paper, motion vectors are estimated in an iterative fashion by means of the Expectation-Maximization (EM) algorithm and a Gaussian data model. Our proposed algorithm also utilizes the local image properties of the scene to improve the motion vector estimates following a spatially adaptive approach. Numerical experiments are presented that demonstrate the merits of our method.

**Keywords:** expectation-maximization; EM; computer vision; motion estimation; pel-recursive, maximum likelihood.

## 1. INTRODUCTION

Motion estimation is very important in multimedia video processing applications. For example, in video coding, the estimated motion is used to reduce the transmission bandwidth. The evolution of an image sequence motion field can also help other image processing tasks in multimedia applications such as analysis, recognition, tracking, restoration, collision avoidance and segmentation of objects [9].

In coding applications, a block-based approach [12] is often used for interpolation of lost information between key frames. The fixed rectangular partitioning of the image used by some block-based approaches often separates visually meaningful image features. If the components of an important feature are assigned different motion vectors, then the interpolated image will suffer from annoying artifacts.

Pel-recursive schemes [2, 4, 5, 9] can theoretically overcome some of the limitations associated with blocks by assigning a unique motion vector to each pixel. Intermediate frames are then constructed by resampling the image at locations determined by linear interpolation of the motion vectors.

The pel-recursive approach can also manage motion with sub-pixel accuracy. However, its original formulation was deterministic. The update of the motion estimate was based on the minimization of the displaced frame difference (DFD) at a pixel. In the absence of additional assumptions about the pixel motion, this estimation problem becomes "ill-posed" because of the following problems: a) occlusion; b) the solution to the 2D motion estimation problem is not unique; and c) the solution does not continuously depend on the data due to the fact that motion estimation is highly sensitive to the presence of observation noise in video images.

In this article, we plan to use the MAP estimate to find $u$, that is, the update of the motion, from our observation model by means of the Espectation-Maximization technique. The main advantage of the EM method is that the final algorithm deals with closed-form expressions and does not require the use of optimization techniques.

We organized this work as follows. Section 2 provides some necessary background on the pel-recursive motion estimation problem. Section 3 introduces our spatially adaptive approach. Section 4 describes the EM framework. Section 5 defines the metric used to evaluate our results. Section 6 describes some implementation aspects and the experiments used to access the performance of our proposed algorithm. Finally, Section 7 has some conclusions.

## 2. PEL-RECURSIVE MOTION ESTIMATION

The displacement of each picture element in each frame forms the displacement vector field (DVF) and its estimation can be done using at least two successive frames. The DVF is the 2D motion resulting from the apparent motion of the image brightness (OF). A vector is assigned to each point in the image.

A pixel belongs to a moving area if its intensity has changed between consecutive frames. Hence, our goal is to find the corresponding intensity value $I_k(r)$ of the $k$-th frame at location $r = [x, y]^T$, and $d(r) = [d_x, d_y]^T$ the corresponding (true) displacement vector (DV) at the working point $r$ in the current frame. Pel-recursive algorithms minimize the DFD function in a small area containing the working point assuming constant image intensity along the motion trajectory. The DFD is defined by

$$\Delta(r; d(r)) = I_k(r) - I_{k-1}(r-d(r)) \quad (1)$$

and the perfect registration of frames will result in $I_k(r) = I_{k-1}(r-d(r))$. The DFD represents the error due to the nonlinear temporal prediction of the intensity field through the DV. The relationship between the DVF and the intensity field is nonlinear. An estimate of $d(r)$, is obtained by directly minimizing $\Delta(r,d(r))$ or by determining a linear relationship between these two variables through some model. This is accomplished by using the Taylor series expansion of $I_{k-1}(r-$

$d(r))$ about the location $(r-d^i(r))$, where $d^i(r)$ represents a prediction of $d(r)$ in $i$-th step. This results in

$$\Delta(r, r-d^i(r)) = -u^T \nabla I_{k-1}(r-d^i(r)) + e(r, d(r)), \quad (2)$$

where the displacement update vector $u = [u_x, u_y]^T = d(r) - d^i(r)$, $e(r, d(r))$ represents the error resulting from the truncation of the higher order terms (linearization error) and $\nabla = [\partial/\partial_x, \partial/\partial_y]^T$ represents the spatial gradient operator. Applying (2) to all points in a neighborhood $\mathcal{R}$ gives

$$z = Gu + n, \quad (3)$$

where the temporal gradients $\Delta(r, r-d^i(r))$ have been stacked to form the $N \times 1$ observation vector $z$ containing DFD information on all the pixels in a neighborhood $\mathcal{R}$, the $N \times 2$ matrix $G$ is obtained by stacking the spatial gradient operators at each observation, and the error terms have formed the $N \times 1$ noise vector $n$ which is assumed Gaussian with $n \sim N(0, \sigma_n^2 I)$. Each row of $G$ has entries $[g_{xi}, g_{yi}]^T$, with $i = 1, \ldots, N$. The spatial gradients of $I_{k-1}$ are calculated through a bilinear interpolation scheme [2].

## 3. SPATIAL ADAPTATION

Aiming to improve the estimates given by the pel-recursive algorithm, we introduced an adaptive scheme for determining the optimal shape of the neighborhood of pixels with the same DV used to generate the overdetermined system of equations given by (3). More specifically, the masks in Fig. 1 show the geometries of the neighborhoods used.

Errors can be caused by the basic underlying assumption of uniform motion inside $\mathcal{R}$ (the smoothness constraint), by not grouping pixels adequately, and by the way gradient vectors are estimated, among other things. Since it is known that in a noiseless image not containing pixels with constant intensity, most errors, when estimating motion, occur close to motion boundaries, we propose a hypothesis testing (HT) approach to determine the best neighborhood shape for a given pixel. We pick up the neighborhood from the finite set of templates shown in Fig. 1, according to the smallest $|DFD|$ criterion, in an attempt to adapt the model to local features associated to motion boundaries.

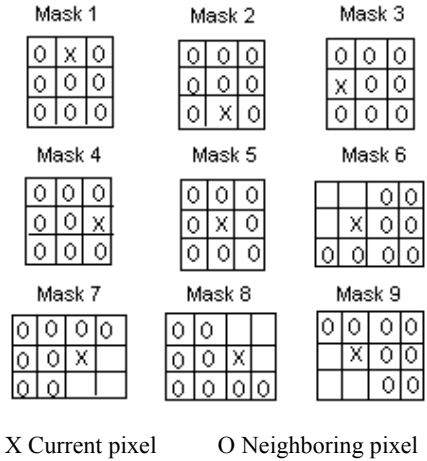

X Current pixel      O Neighboring pixel

Figure 1: Neighborhood geometries.

## 4. THE EM APPROACH

### 4.1. The Ordinary ML Estimate and Its Limitations

The MAP estimate of $u$ is given by [11]

$$\hat{u}_{MAP} = \Lambda_u G^T (G \Lambda_u G^T + \Lambda_n)^{-1} z. \quad (4)$$

The computation of $\hat{u}_{MAP}$ requires knowledge of the covariance matrices of the update vector ($\Lambda_u$) and the noise/linearization error term ($\Lambda_n$), respectively.

The estimate in Eq. (4) was derived assuming that u and n have zero mean, and are uncorrelated. The MAP for the linear observation model in Eq. (3) is also the maximum a posteriori (MAP) estimate, assuming a Gaussian prior on $u$ and Gaussian noise $n$ (for more details, see [11]).

In this work, we resort to the maximum likelihood (ML) estimation of the second-order statistics $\Lambda_u$ and $\Lambda_n$ of the model. The calculation of the ML estimates of $\Lambda_u$ and $\Lambda_n$ is done iteratively by means of the Expectation-Maximization (EM) algorithm formalized by Dempster et al [3] which has been used in a variety of applications [7, 8, 10]. The EM algorithm was previously used for motion estimation in [6]. However, this work deals with estimate parameters of affine models and it uses a block-matching framework.

For the model described by Eq. (3), let us assume that the update vector $u$ and the noise $n$ are normally-distributed with mean zero and covariance matrices equal to $\Lambda_u$ and $\Lambda_n$, respectively. If we consider $u$ and $n$ as being uncorrelated, then the pdf of $z$ is

$$f_z(z) = \left| 2\pi (G \Lambda_u G^H + \Lambda_n) \right|^{-\frac{1}{2}} \times exp \left\{ -\frac{1}{2} z^H (G \Lambda_u G^H + \Lambda_n)^{-1} z \right\}.$$

Let us take $\Phi = \{\Lambda_u, \Lambda_n\}$ as the parameter set to be estimated. The pdf of $z$ can be considered a continuous and differentiable function of $\Phi$. This dependency can be better captured by the notation $f_z(z; \Phi)$. Furthermore, we can assume that the additive noise is white, with covariance matrix $\Lambda_n = \sigma_n^2 I$, where $I$ is an $N \times N$ identity matrix.

The ML estimation of $\Phi$ is the $\Phi_{ML}$ that maximizes the logarithm of the likelihood function of $f_z(z; \Phi)$, which is given by

$$\Phi_{ML} = arg\{ \max_{\Phi} f_z(z; \Phi) \} = arg\{ \max_{\Phi} \log f_z(z; \Phi) \}. \quad (5)$$

Combining Eqs. (4) and (5), we conclude that the maximization of the log-likelihood function is equivalent to minimizing the function $L_o(\Phi)$ where

$$L_o(\Phi) = \log \left| G \Lambda_u G^H + \Lambda_n \right| + z^H (G \Lambda_u G^H + \Lambda_n)^{-1} z.$$

$L_o(\Phi)$ is a nonlinear and non-unimodal function with respect to $\Phi$. Analytical solutions for the previous equation are out of question. Thus, we have to resort to optimization techniques. Nevertheless, since $L_o(\Phi)$ is not unimodal, convergence and local extrema of iterative optimization methods are serious problems. The EM algorithm arises as an alternative for estimating $\Phi$, since its convergence to a local maximum is guaranteed.

The EM algorithm is a general numerical technique which can be used to determine the maximum likelihood estimate (MLE) of a set of parameters. It can be employed for identification of model and distribution parameters simultaneously. The parameters can be estimated iteratively even in situations where some variables cannot be observed.

### 4.2. Problem Formulation According to the EM Framework

We assume that the update vector $u$ normally-distributed with mean zero and covariance matrix equal to $\Lambda_u$, and $n \sim N(0, \sigma_n^2 I)$ is the additive Gaussian noise.

If we choose $x = [u \;\; n]^T$ as the complete data, where $x \in \mathbb{R}^{(N+2)\times 1}$, then $x$ will be normally distributed with diagonal covariance matrix $\Lambda_x$ equal to

$$\Lambda_x = \begin{bmatrix} \Lambda_u & 0 \\ 0 & \sigma_n^2 I \end{bmatrix} = \begin{bmatrix} \Lambda_u & 0 \\ 0 & \Lambda_n \end{bmatrix},$$

with its corresponding inverse

$$\Lambda_x^{-1} = \begin{bmatrix} \Lambda_u^{-1} & 0 \\ 0 & \Lambda_n^{-1} \end{bmatrix}.$$

Eq. (3) can be re-written as

$$z = [G; I]\begin{pmatrix} u \\ n \end{pmatrix} = Hx.$$

The foundation of the EM algorithm is the maximization of the expectation of $\log\{f_x(x;\Phi)\}$ given the incomplete observed data $z$ and the current estimate of the parameter set $\Phi$. $f_x(x;\Phi)$ is the pdf of the complete data and it is given by

$$f_x(x;\Phi) = |2\pi \Lambda_x|^{-\frac{1}{2}} \exp\left\{-\frac{1}{2} x^H \Lambda_x^{-1} x\right\}. \tag{6}$$

Taking the logarithm of both sides of Eq. (11) leads to

$$\log\{f_x(x;\Phi)\} = -\frac{1}{2}\log|2\pi\Lambda_x| - \frac{1}{2} x^H \Lambda_x^{-1} x$$
$$= K - \frac{1}{2}[\Lambda_x + x^H \Lambda].$$

where $\Phi = \Lambda_x$ and $K$ is a constant independent of $\Phi$. The E-step requires the computation of the expectation

$$Q(\Phi;\Phi^{(p)}) = E[\log\{f_x(x;\Phi)\} | z; \Phi^{(p)}],$$

where $\Phi^{(p)} = [\Lambda_u^{(p)}; \Lambda_n^{(p)}]$ is the estimate of the parameter set $\Phi = \{\Lambda_u; \Lambda_n\}$ at the $p$-th iteration. The M-step maximizes $Q(\Phi;\Phi^{(p)})$:

$$\Phi^{(p+1)} = \arg\{\max_\Phi Q(\Phi;\Phi^{(p)})\}.$$

The E-step can be re-written in terms of the parameter set $\Phi$ by means of the relationships developed in terms of the complete data $x$ and the pdf $f(x|z;\Phi^{(p)})$ as

$$F(\Phi;\Phi^{(p)}) = \log|\Lambda_u| + \log|\Lambda_n| + tr(\Lambda_u^{-1}\Lambda_{u|z}^{(p)})$$
$$+ tr(\Lambda_n^{-1}\Lambda_{n|z}^{(p)}) + \mu_{u|z}^{(p)H} \Lambda_u^{-1} \mu_{u|z}^{(p)} + \mu_{n|z}^{(p)H} \Lambda_n^{-1} \mu_{n|z}^{(p)}.$$

Our goal is to estimate the update vector $u$. The MAP/MMSE estimate of $u$ is $\mu_{u|z}^{(p)}$ which corresponds to its conditional expectation at iteration p, given the observation $z$. Since our observation model is linear and Gaussian statistics are assumed, we have

$$\hat{u}_{MMSE} = \hat{u}_{LMMSE} = \hat{u}_{MAP}.$$

This estimate is obtained as a byproduct of the estimation of $\Phi$. The statistical assumptions about $u$ and $n$ result in $\Phi = \{\sigma_1^2, \sigma_2^2, \sigma_n^2\}$. The two steps of the resulting EM algorithm are

**E-Step**:

$$F(\Phi;\Phi^{(p)}) = \log(\sigma_1^2) + \log(\sigma_2^2)$$
$$+ m[\log(\sigma_n^2)] + \frac{a_{11}^{(p)} + c_1^{2(p)}}{\sigma_1^2} + \frac{a_{22}^{(p)} + c_2^{2(p)}}{\sigma_2^2}$$
$$+ \frac{1}{\sigma_n^2}[b_{11}^{(p)} + \cdots + b_{mm}^{(p)} + e_1^{2(p)} + \cdots + e_m^{2(p)}],$$

where

$$\mathbf{A}^{(p)} = \begin{bmatrix} a_{11}^{(p)} & a_{12}^{(p)} \\ a_{21}^{(p)} & a_{22}^{(p)} \end{bmatrix} = \left[\Lambda_u^{(p)} - \Lambda_u^{(p)} \mathbf{G}^H (\mathbf{G}\Lambda_u^{(p)}\mathbf{G}^H + \Lambda_n^{(p)})^{-1} \mathbf{G}\Lambda_u^{(p)}\right],$$

$$\mathbf{B}^{(p)} = \left[\Lambda_n^{(p)} - \Lambda_n^{(p)}\left(\mathbf{G}\Lambda_u^{(p)}\mathbf{G}^H + \Lambda_n^{(p)}\right)^{-1} \Lambda_n^{(p)}\right]$$
$$= \begin{bmatrix} b_{11}^{(p)} & b_{12}^{(p)} & \cdots & b_{1m}^{(p)} \\ b_{21}^{(p)} & b_{22}^{(p)} & \cdots & b_{2m}^{(p)} \\ \vdots & \vdots & \ddots & \vdots \\ b_{m1}^{(p)} & b_{m2}^{(p)} & \cdots & b_{mm}^{(p)} \end{bmatrix},$$

$$c^{(p)} = \mu_{u|z}^{(p)} = \begin{bmatrix} c_1^{(p)} \\ c_2^{(p)} \end{bmatrix} = \left[\Lambda_u^{(p)} \mathbf{G}^H (\mathbf{G}\Lambda_u^{(p)}\mathbf{G}^H + \Lambda_n^{(p)})^{-1} z\right], \quad (7)$$

and $\quad e^{(p)} = \mu_{n|z}^{(p)} = \begin{bmatrix} e_1^{(p)} \\ \vdots \\ e_m^{(p)} \end{bmatrix} = \left[\Lambda_n^{(p)}(\mathbf{G}\Lambda_u^{(p)}\mathbf{G}^H + \Lambda_n^{(p)})^{-1} z\right].$

**M-Step:**

In order to obtain this step, we need to find the minimum of $F(\Phi,\Phi^{(p)})$. Differentiating $F(\Phi,\Phi^{(p)})$ with respect to $\sigma_n^2$ and making it equal to zero yields

$$\sigma_n^{2(p+1)} = \frac{1}{m}\left\{Tr\left[\mathbf{B}^{(p)}\right] + \left|e^{(p)}\right|^2\right\}. \tag{8}$$

Applying similar procedure for $\sigma_1^2$ and $\sigma_2^2$, gives

$$\sigma_1^{2(p+1)} = a_{11}^{(p)} + c_1^{2(p)}, \text{ and} \tag{9}$$
$$\sigma_2^{2(p+1)} = a_{22}^{(p)} + c_2^{2(p)}. \tag{10}$$

Now, we are ready to state the resulting EM algorithm using multiple masks.

### 4.3. The EM-Based Motion Estimation Descrptive Algorithm

For each pixel in the current frame $k$, located at $r$, do the following:

1) Initialize the system: $d^0(r), \sigma_n^{2(0)}, \sigma_1^{2(0)},$ and $\sigma_2^{2(0)}$, $m \leftarrow 0$ ($m$=mask counter), and $p \leftarrow 0$ ($p$= iteration counter).

2) If $|DFD| < T$, then stop. $T$= threshold for $|DFD|$.
3) Calculate $G^i$ and $z^i$ for the current mask and current initial estimate.
4) Calculate the current update vector by means of Eq. (7), (byproduct of the E-Step).
5) Perform the M-Step and update the parameter estimates using Eqs. (8), (9), and (10).
6) Calculate the new displacement vector:
$$d^{p+1}(r) = d^p(r) + u^p. \qquad (11)$$
7) For the current mask m:
If $\|\Phi^{p+1} - \Phi^p\| \leq \xi$ (convergence test for the EM algorithm), $\|d^{p+1}(r) - d^p(r)\| \leq \varepsilon$ and $|DFD| < T$, then stop.

Otherwise, if $p < (I-1)$, where $I$ is the maximum number of iterations allowed, then go to step 3 with $p \leftarrow p+1$.

If $p=I-1$, try another neighborhood geometry $(m \leftarrow m+1)$, and reset variables: $p \leftarrow 0$, $d^0(r)$, $\sigma_n^{2(0)}$, $\sigma_1^{2(0)}$, and $\sigma_2^{2(0)}$. Go to step 2.

If all masks where used and no displacement vector was found, then set $d^{p+1}(r) = 0$.

## 5. METRIC

This work assesses the motion field quality through the use of the fowlling metric [2, 4, 5]:

### 5.1 Improvement in Motion Compensation

The $\overline{IMC(dB)}$ between two consecutive frames is given by

$$\overline{IMC}_k(dB) = 10\log_{10}\left\{\frac{\sum_{r \in S}[I_k(r) - I_{k-1}(r)]^2}{\sum_{r \in S}[I_k(r) - I_{k-1}(r - d(r))]^2}\right\}, \text{ where S is}$$

the frame being currently analyzed. It shows the ratio in decibel ($dB$) between the mean-squared frame difference ($\overline{FD}^2$) defined by

$$\overline{FD}^2 = \frac{\sum_{r \in S}[I_k(r) - I_{k-1}(r)]^2}{RC}$$

and the $\overline{DFD}^2$ between frames $k$ and $(k-1)$.

As far as the use of the this metric goes, we chose to apply it to a sequence of $K$ frames, resulting in the following equation for the average improvement in motion compensation:

$$\overline{IMC}(dB) = 10\log_{10}\left\{\frac{\sum_{k=2}^K \sum_{r \in S}[I_k(r) - I_{k-1}(r)]^2}{\sum_{k=2}^K \sum_{r \in S}[I_k(r) - I_{k-1}(r - d(r))]^2}\right\}$$

When it comes to motion estimation, we seek algorithms that have high values of $\overline{IMC}(dB)$. If we could detect motion without any error, then the denominator of the previous expression would be zero (perfect registration of motion) and we would have $\overline{IMC}(dB) = \infty$.

## 6. EXPERIMENTS

This section presents results illustrating the effectiveness of the EM algorithm and compare it to the Wiener filter described by

$$\hat{u}_{Wiener} = \hat{u}_{LMMSE} = (G^T G + \mu I)^{-1} G^T z,$$

with $\mu=50$ and constant for the entire frame [1]. The algorithms were tested using two real video sequences: the "Foreman" and the "Mother and Daughter" (MD). The sequences are 144 x 176, 8-bit (QCIF). The algorithm called "EM" uses one mask, while algorithm "EMm" employs multiple masks.

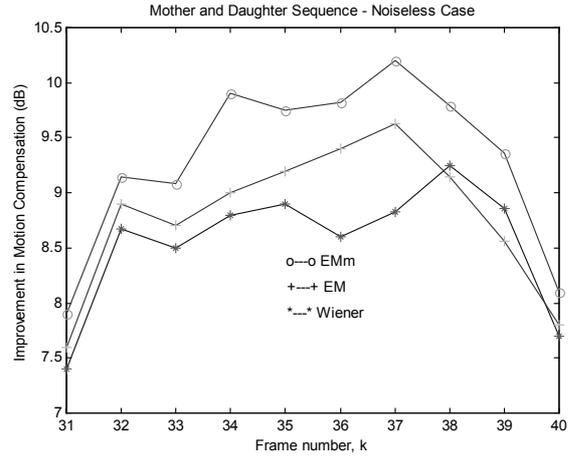

(a)

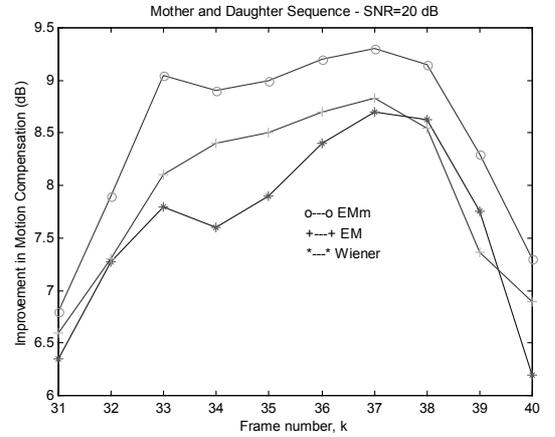

(b)

Figure2: $\overline{IMC(dB)}$ for the noiseless (a) and noisy (b) cases for the "MD" sequence.

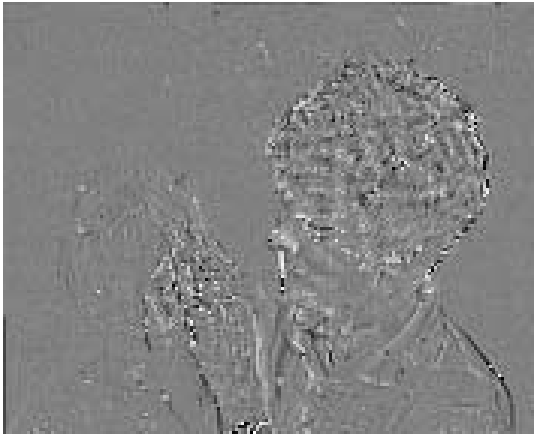

(a)

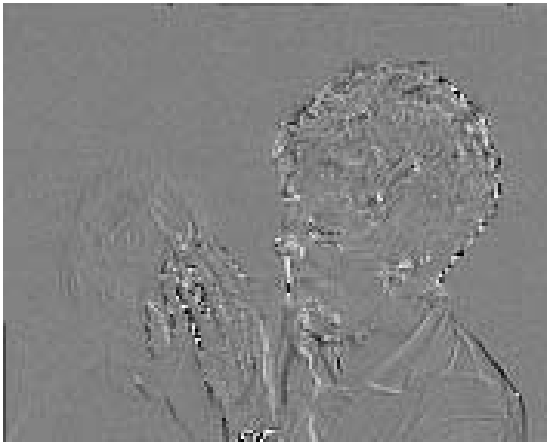

(b)

Figure 3: Motion-compensated errors for frame 32 of the "Mother and Daughter" sequence with SNR=20dB: the Wiener (a) and the EMA (b) algorithms.

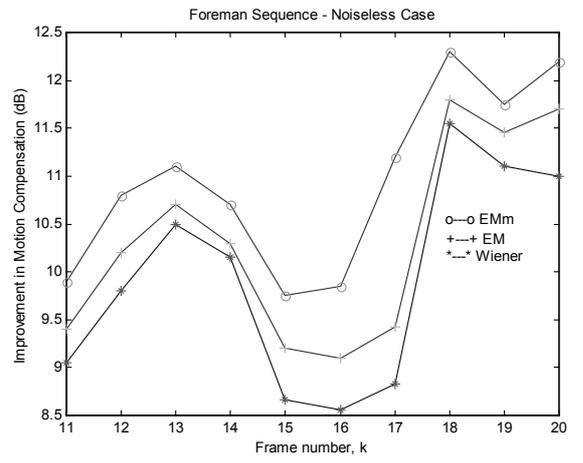

(a)

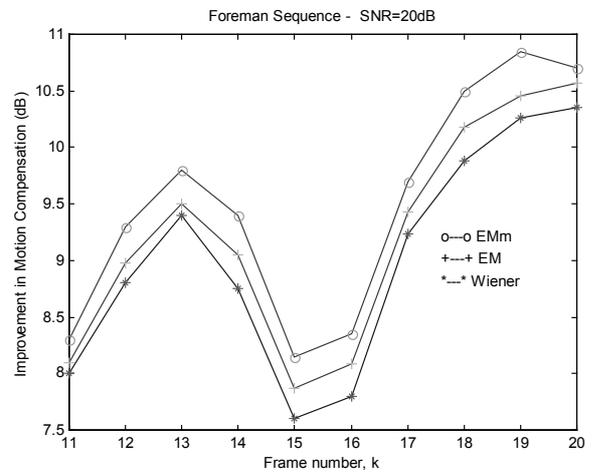

(b)

Figure 4: $\overline{IMC}(dB)$ for frames 11-20 of the noiseless (a) and noisy cases with SNR=20dB (b) for the "Foreman" sequence.

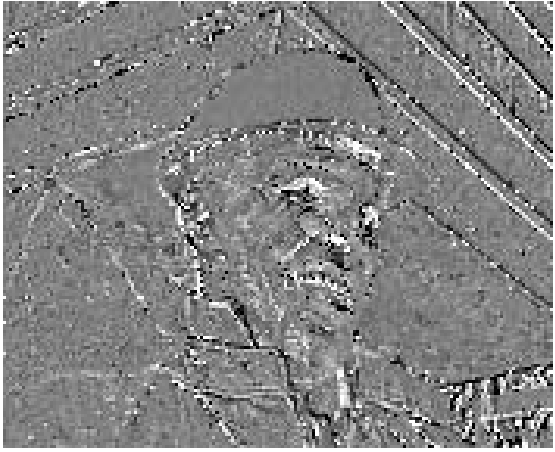

(a)

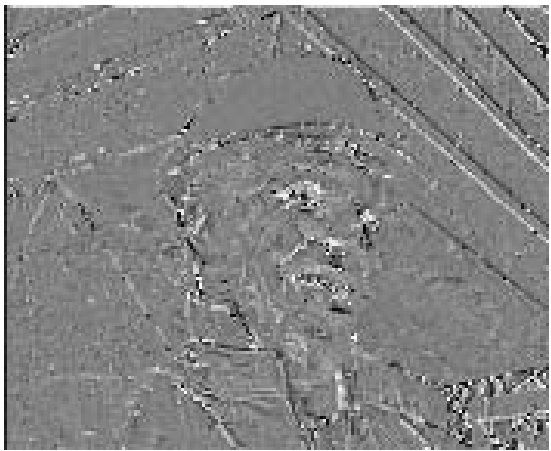

(b)

Figure 5: Motion-compensated errors for frame 16 of the "Foreman" sequence with SNR=20dB: the Wiener (a) and the EMA (b) algorithms.

## 7. CONCLUSIONS

For all the sequences, the EM algorithm performed better than the Wiener filter for both the one mask and multi-mask cases, regardless of the presence of noise.

The EM algorithm showed some sensitivity to the choice of initial estimates. We used more than one initial parameter set $\boldsymbol{\Phi}^0 = [\sigma_n^{2(0)}, \sigma_1^{2(0)}, \sigma_2^{2(0)}]^T$ to improve the rate of convergence, but even with this extra feature, the resulting algorithms using one mask and multiple masks where faster than the corresponding Wiener filter counterparts.

For the EM method, we have a simple algorithm that is guaranteed to converge and it does not equire numerical optimization. Given that there are multiple iterations at every pixel location, the speed advantage gained by means of the EM algorithm is considerable. The authors think this framework has great potential for applications such as video coding and image segmentation.